\documentclass[a4paper, 10 pt, conference]{hsmr} 
\usepackage[utf8]{inputenc}
\IEEEoverridecommandlockouts                       
\usepackage{mathtools}
\usepackage{geometry}
\usepackage{gensymb}
\usepackage[labelsep=space]{caption}
\usepackage{newtxmath,newtxtext}
\usepackage{authblk}
\usepackage{pgfplots}
\usepackage{graphicx}
\usepackage[colorlinks,urlcolor=blue]{hyperref} 
\usepackage{newtxtext,newtxmath}

\pgfplotsset{compat=newest} 
 
\title{\LARGE \bf
Incident Angle Study for Designing an Endoscopic Tool \\for Intraoperative Brain Tumor Detection}

\author{\LARGE K. K. Yamamoto$^{1}$}
\author{\LARGE T. J. Zachem$^{1,2}$}
\author{\LARGE W. Ross$^{2}$} 
\author{\LARGE P. J. Codd$^{1,2}$}

\affil{\Large\textit{$^{1}$Department of Mechanical Engineering and Materials Science, Duke University}\\ \Large\textit{$^{2}$Department of Neurosurgery, Duke University School of Medicine}\\ \Large\textit{kent.yamamoto@duke.edu}}

\begin{document}

\maketitle
\thispagestyle{empty}
\pagestyle{empty}

\section*{INTRODUCTION}

Surgery is one of the most prevalent methods of controlling and eradicating \textcolor{black}{tumor} growth in the human body, with a projection of 45 million surgical procedures per year by 2030 \cite{surgeryStat}. In brain tumor resection surgeries, preoperative images used for the detection and localization of the cancer regions become less reliable throughout surgery when used intraoperatively due to the brain moving during the procedure, referred to as brain shift. To solve the brain shift problem, intraoperative MRI (iMRI) has been used, but it is costly, time intensive, and only available at the most advanced care facilities \cite{imri}. Intraoperative fluorescence-guided methods, both exogenous (introducing foreign fluorophore molecules into  the body) and endogenous (utilizing innate fluorophores within the body), have been investigated as an alternative to iMRI. 

This paper introduces the proposed design, shown in Fig. 1(a), of an endoscopic tool for intraoperative brain tumor detection and characterization, incorporating a laser-based endogenous fluorescence method previously explored by \cite{tucker1}, called TumorID, depicted in Fig. 1(b). The device has also been deployed on ex-vivo pituitary adenoma tissue by \cite{zachem} for \textcolor{black}{intraoperative} \textcolor{black}{pituitary adenoma identification and subtype classification}. This study explores whether a non-perpendicular angle of incidence (AoI) will significantly affect the emitted spectral data. With a better understanding of the relationship between AoI and collected spectra, the results can help shed light on the potential steering modality $($optical \cite{wood} or fiber \cite{hybrid}$)$ and end-effector movement profile for the proposed optics-based endoscopic tool. 

\section*{MATERIALS AND METHODS}
The experimental setup, Fig. 2(a), consisted of TumorID and its components, in addition to a spectrally-tuned gellan gum phantom model. The objective of the study was to scan the phantom model at different incident angles to observe the effects, if any, in the emitted spectral data. 

\begin{figure}[h!]
    \centering
    \includegraphics[scale = 0.29]{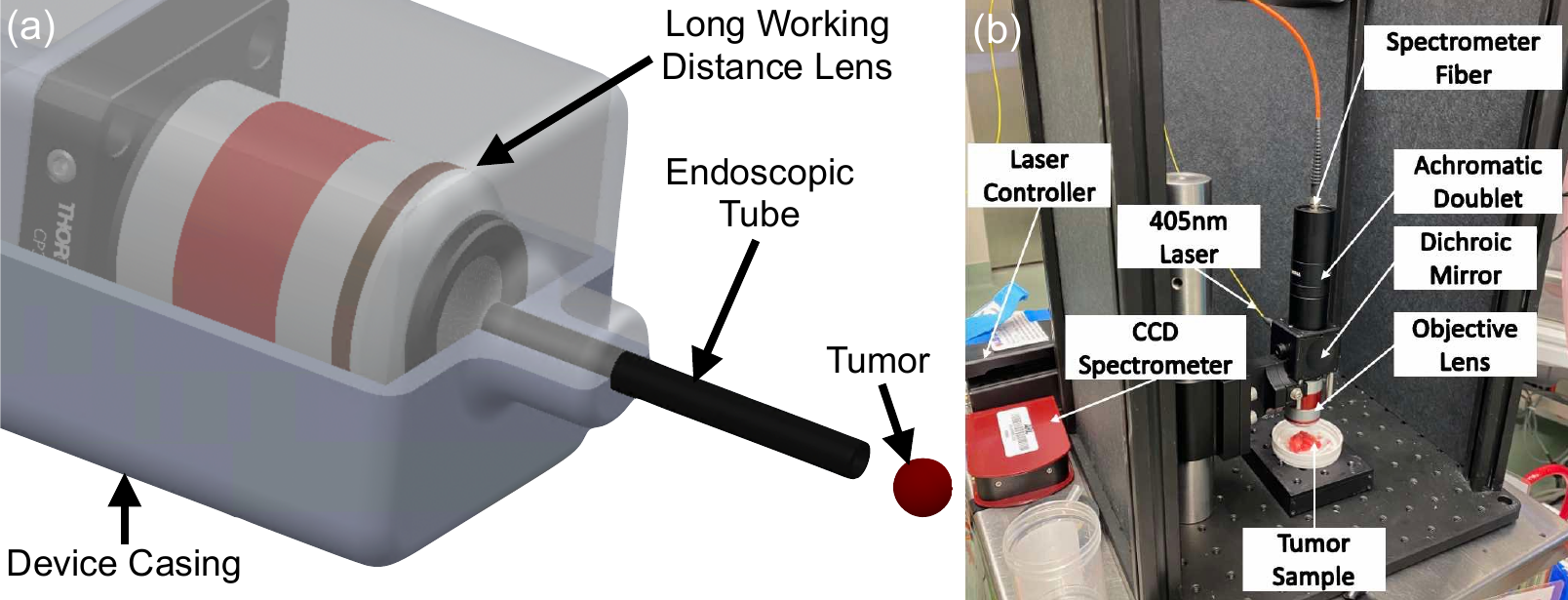}
    \caption{\small a) Proposed endoscopic tumor detection device. b) Current TumorID setup scanning a tumor sample.}
    \label{fig:intro}
    \vspace{-0.5cm}
\end{figure}

\subsection{TumorID}
The TumorID, a non-contact tumor detection device completed in previous work by \cite{tucker1}, leveraged the fact that NADH and FAD are endogenous fluorophores to classify tissue type based on spectral signatures. The spectral signatures between tumor and healthy tissue differed due to the Warburg Effect, resulting in different concentrations in healthy and tumor cells due to differing metabolic properties. The system used a 405 nm laser diode (Thorlabs, NJ, USA) to irradiate the tissue. The emitted light was guided through a collimating lens to a dichroic mirror, then focused on the sample through an objective lens.

The TumorID was attached to a NEMA 17 hybrid stepper motor (Moon's, Shanghai, China) using a standard optical post for rigid attachment, as shown in Fig. 2(b). The stepper motor was controlled using a microcontroller (Elegoo, Shenzhen, China) and motor controller (SparkFun Electronics, CO, USA). The motor controller was powered by an external power supply at 11V. Spectral data were collected by a Thorlabs CCS200 spectrometer (Thorlabs, NJ, USA).

\begin{figure}[h!]
    \centering
    \includegraphics[scale = 0.39]{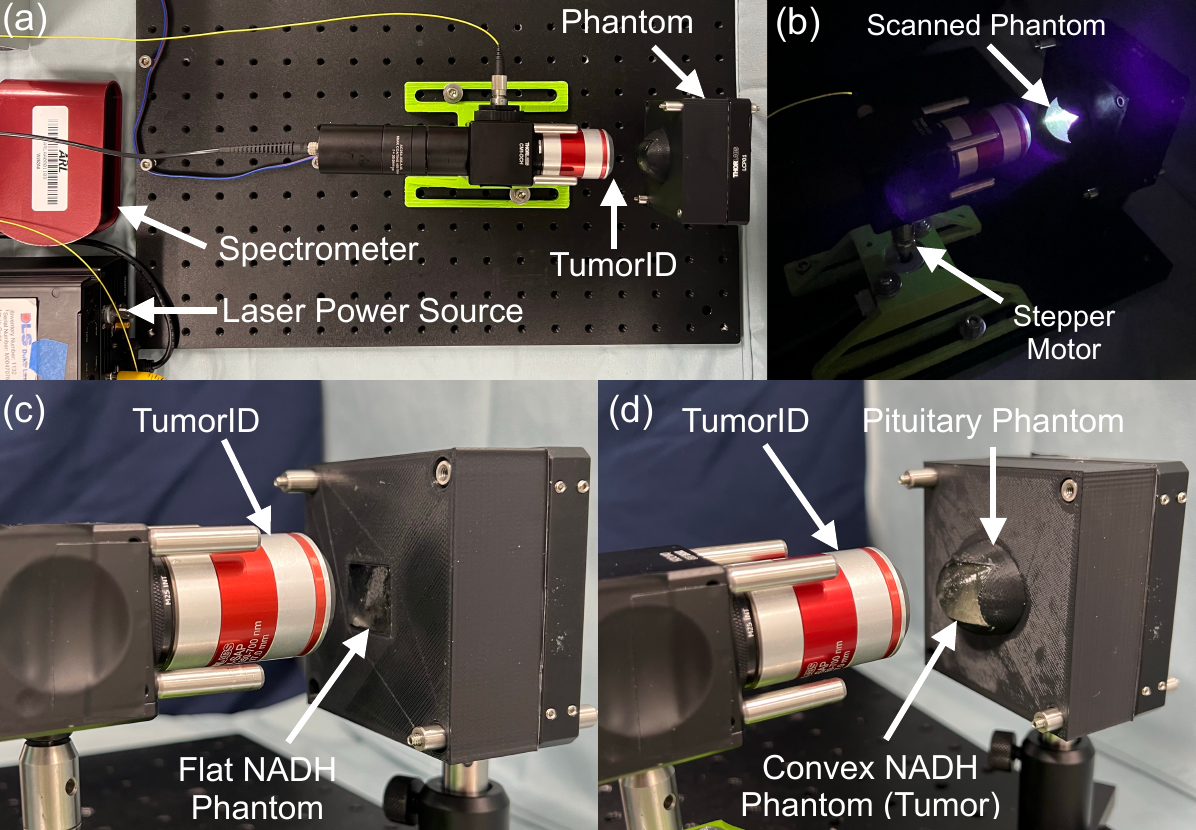}
    \caption{\small (a) Experimental setup for TumorID AoI study. (b) TumorID mounted on a stepper motor, scanning the convex phantom. (c) Flat phantom (control) prepared for angle of incidence study. (d) Convex phantom (representing a pituitary tumor) fixed in skull base phantom frame.}
    \label{fig:methods}
    \vspace{-0.3cm}
\end{figure}

\subsection{Phantom Model}
Two gellan-gum phantom models with varying geometries (a flat plate and a spherical tumor representing a pituitary tumor) were prepared using NADH and FAD to mimic the spectral properties of tumors. The concentrations and volumes of both NADH and FAD were used from previous work on creating tumor-mimicking phantom models \cite{phantom}. The phantom solution was cured in both a flat and convex mold that was 3D printed (UltiMaker S3, Utrecht, Netherlands). The cured phantoms were then placed in their respective fixtures, with the convex fixture representing a pituitary tumor's geometry during an endonasal approach. Both flat and convex phantoms are shown in the experimental setup in Fig. 2(c) and Fig. 2(d), respectively.

\subsection{Experimental Design}
The rotating TumorID was placed 17 mm, the objective lens' working distance, from the phantom at an incident angle of 90$\degree$. To initiate the experiment, the TumorID was placed at the edge of the phantom boundary, then swept clockwise in increments of $1.8 \degree$ to collect a spectral signal at each incident angle ($18 \degree$ in both directions with respect to an AoI of 0$\degree$). 

The procedure was executed on both the flat phantom and the curved phantom in triplicates. The Area Under the Curve (AUC) of each spectrum was calculated as a representative metric for the study to observe whether the spectral signal significantly changed at each incident angle for both phantoms. After spectral collection, the data was normalized individually with respect to the largest intensity value greater than 450 nm \cite{zachem}. 450 nm was selected as the cutoff for normalization due to the transmittance behavior of the dichroic mirror. After max normalization, the data was smoothed by convolving with a window size of two. The AUC was calculated by conducting trapezoidal integration of the normalized signal from 450 nm to 750 nm. For clarity of presentation, AUC for each phantom model was normalized via max normalization.

\section*{RESULTS}
The spectra at each AoI were recorded for both the flat and convex phantom, the latter shown in Fig. 3(a). Each line represented the spectra observed at an incident angle. Spectra collected at each incident angle resulted in the emission peaks for NADH and FAD \cite{tucker1}, found at 460 nm and 525 nm, respectively. Fig. 3(b) showed the relationship between normalized AUC values and incident angle for both the flat and convex phantoms. The AUC values within $95 \%$ of the max value resided within $\pm$ $18\degree$ from the zero position for the flat phantoms, decreasing as AoI increases. The AUC values calculated for the convex phantom showed a similar trend, but the values within $95 \%$ of the max value were present in a smaller region ($\pm$ $14 \degree$). The mean AUC for the flat phantom was 0.98 $\pm$ 0.01, and the mean AUC for the convex phantom was 0.98 $\pm$ 0.02.

\begin{figure}[h!]
    \centering
    \includegraphics[scale = 0.29]{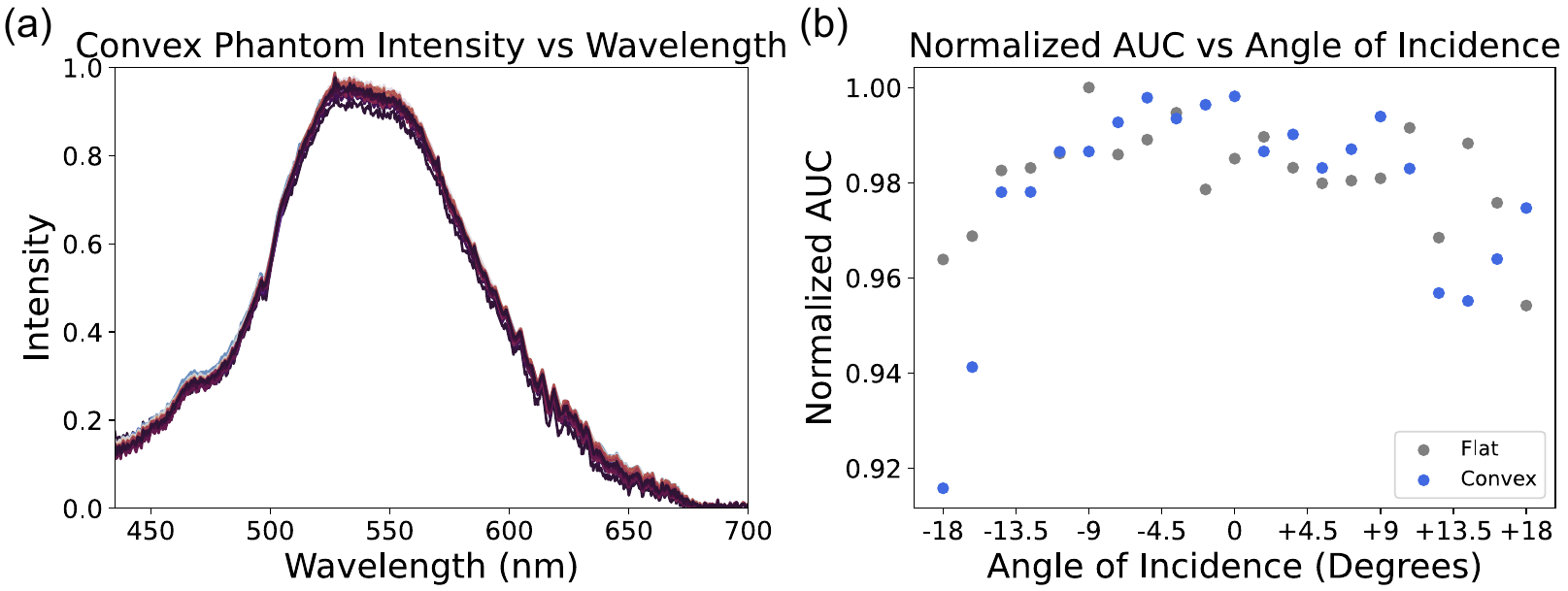}
    \caption{\small (a) Spectral responses of the convex phantom at each incident angle, n = 21. (b) Average normalized AUC values at tested AoI.}
    \label{fig:results}
\end{figure}

\section*{DISCUSSION}
A study on a non-contact tumor identification device, the TumorID, was conducted to investigate the effect of AoI on spectral response. The data collected for both a flat and convex scanning surface showed that although there was a change in spectra intensity at various AoI's, there was negligible difference in the spectral signature at the 460 nm and 525 nm wavelengths, where NADH and FAD emissions were present. The AUC values calculated were all within an average of 0.98 and \textcolor{black}{a standard deviation of 0.02}, indicating that the change in spectra intensity due to AoI was also very minimal after normalization. These findings were expected but also helped confirm that the end-effector of the proposed endoscopic device may not need to be continuously perpendicular to the scanning surface. Future work will include designing and implementing a compact steering method for bidirectional optics toward minimally invasive intraoperative tumor detection.

\nocite{*}
\bibliographystyle{IEEEtran}
\bibliography{HSMR}

\end{document}